%% file: main.tex
\documentclass{article}
\usepackage{spconf,amsmath,graphicx,hyperref}
\usepackage{booktabs}
\hypersetup{hidelinks}

\newif\ifpaperhasreferences
\paperhasreferencestrue

\title{LEARNING NATURAL CONVERSATIONAL BEHAVIOR IN TANDEM SPEECH-TO-SPEECH MODELS WITH RANDOMIZED GUIDANCE}

\name{Manato Yaguchi$^{1,2}$, \qquad Yotaro Kubo$^{1}$, \qquad Hikaru Asano$^{1,2}$, \qquad So Kuroki$^{1}$}
\address{$^{1}$Sakana AI, Tokyo, Japan\thanks{This work was done during Manato Yaguchi's internship at Sakana AI.} \qquad
$^{2}$The University of Tokyo, Tokyo, Japan}

\begin{document}

\maketitle

\begin{abstract}
\input{sections/abstract}
\end{abstract}


\input{sections/introduction}
\input{sections/method}
\input{sections/experiments}
\input{sections/conclusion}

\ifpaperhasreferences
  \bibliographystyle{IEEEbib}
  \bibliography{refs}
\fi

\end{document}

%% file: sections/abstract.tex
Tandem speech-to-speech architectures couple a responsive speech frontend with an asynchronous text backend. 
In KAME, a large language model (LLM) serves as the backend, supplying candidate responses as guidance to the speech frontend while the user is still speaking. 
Ordinary conversation recordings capture the eventual response but not the guidance the backend would supply during the user's utterance.
Generating the missing guidance with a simulator LLM adds substantial data-preparation overhead when training on real conversations. 
We propose randomized intermediate guidance, which derives guidance directly from the conversation corpus rather than simulating backend LLM behavior. 
During training, target responses provide informative guidance, while randomly sampled responses provide potentially irrelevant updates during the utterance. This combination aims to teach the frontend to use backend information selectively.
On synthetic dialogues, KAME trained with this recipe achieves response quality comparable to that of the LLM-generated and similarity-based baselines. 
Training on 3.8k hours of real conversations improves smooth turn-taking and audio-judge naturalness over synthetic-data KAME while retaining a response-quality advantage over Moshi.
These results show that randomized guidance offers a practical route to combining the response-quality benefits of tandem models with natural conversational behavior learned from real speech.

%% file: sections/introduction.tex
\section{Introduction}
\label{sec:introduction}

A central challenge in real-time spoken dialogue is to combine high-quality
responses with natural interaction. Full-duplex speech-to-speech (S2S) models
such as Moshi support low-latency speech generation and flexible turn-taking
\cite{defossez2024moshi}, but their knowledge and reasoning capabilities remain
limited. Cascaded systems can draw on powerful text-based large language
models (LLMs), but their sequential ASR--LLM--TTS pipeline introduces delays
that can disrupt conversational flow. 

To combine the strengths of both approaches, tandem architectures couple a responsive speech frontend with
an asynchronous backend.
KAME exemplifies this design by guiding a full-duplex S2S frontend with
responses from an asynchronous text LLM \cite{kuroki2026kame}. 
MoshiRAG retrieves external knowledge during full-duplex speech
generation \cite{chien2026moshirag}, while ConvFill lets a lightweight Talker
respond as a slower Reasoner streams knowledge \cite{srinivas2026thinking}.
For such asynchronous interfaces, a desirable capability is to use relevant backend information without relying on irrelevant updates, while maintaining natural conversational timing and delivery.

To train such tandem S2S systems, supervised fine-tuning requires not only ordinary dialogue pairs but also the intermediate guidance supplied by the backend.
This makes it difficult to use transcribed dialogue audio as a training dataset for the tandem architecture, because such datasets lack the intermediate guidance supplied by the backend LLM to the speech frontend.
To mitigate this issue, KAME, for example, employs guidance generated by a simulator LLM \cite{kuroki2026kame}. 
The simulator LLM is instructed to generate provisional responses periodically while a user utterance is ongoing. 
The simulation process is designed so that the generated responses gradually become semantically closer to the target response, i.e., what the next speaker actually said, as more of the user’s utterance is observed.
This design is intended to supply increasingly useful guidance as the user utterance unfolds, encouraging the frontend to exploit informative backend updates during response generation.

This per-example LLM simulation becomes a data-preparation bottleneck when scaling tandem training to large real-conversation corpora, limiting the use of real speech for learning natural conversational behavior.
The original KAME is trained on synthetic dialogues derived from question--answer pairs and converted to speech \cite{kuroki2026kame}. These data improve response quality but do not directly capture the timing and vocal delivery of spontaneous conversation. 
Recent work has explored real conversations for interactivity
alignment \cite{ohashi2026interactivity} and full-duplex data construction
\cite{he2026conversationalvoice,nakata2026duplexchat}. However, these efforts focus on full-duplex modeling or data construction rather than scaling real-conversation training in tandem architectures.

To eliminate this bottleneck, we introduce randomized intermediate guidance. 
Instead of faithfully simulating the backend's intermediate guidance, our proposed method uses a sequence of informative and potentially irrelevant guidance texts.
Following KAME, the target response is supplied as informative guidance, and randomly sampled response texts from other conversations are supplied as potentially irrelevant guidance in place of KAME's intermediate guidance.
By mixing those two types of guidance, we aim to encourage the frontend model to distinguish and exploit useful information.
This allows all the training data to be constructed directly from conversational data with transcripts, without per-example LLM-based simulation.

We evaluate this approach in two complementary experiments on the KAME architecture. 
First, on synthetic dialogues, randomized guidance is shown to be competitive with LLM-generated guidance and similarity-based retrieval on spoken MT-Bench, whereas Target-only training yields lower scores (Table~\ref{tab:synthetic-guidance}). 
We then apply randomized guidance to train KAME on a 3.8k-hour corpus of real two-speaker conversations. 
The resulting model achieves smoother turn-taking and higher audio-judge naturalness than synthetic-data KAME, while retaining an MT-Bench score of 5.04 versus Moshi's 1.96 (Table~\ref{tab:natural-conversations}). 
Together, these results demonstrate a practical way to combine the response-quality benefits of tandem S2S with natural interaction learned from real conversations.

%% file: sections/method.tex
\begin{figure*}[t]
    \centering
    \includegraphics[width=1.0\textwidth]{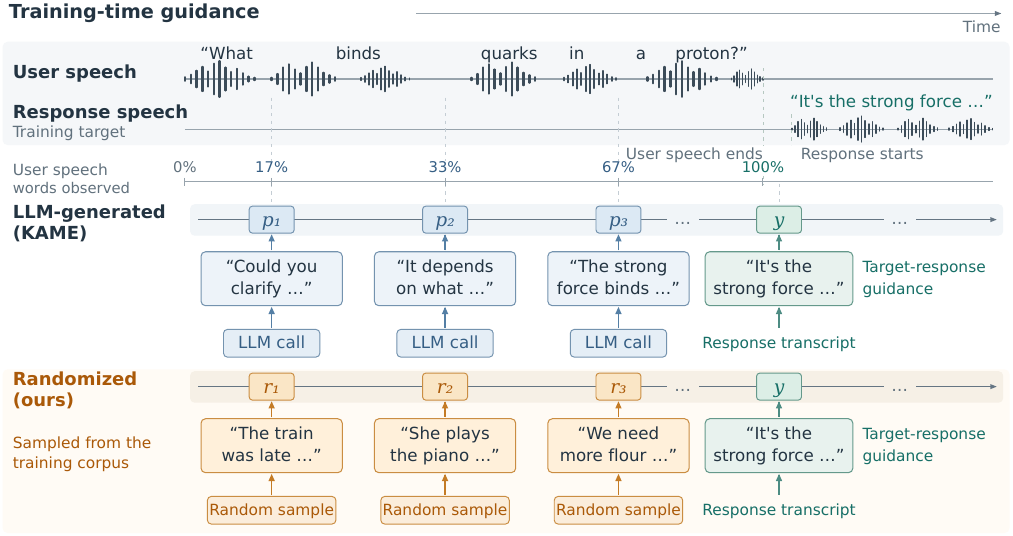}
    \caption{
    Schematic training-time guidance streams: $p_i$, simulator-LLM-generated responses; $r_i$, randomly sampled responses; $y$, the target transcript. The inference-time backend is unchanged.
    }
    \label{fig:guidance-overview}
\end{figure*}

\section{Randomized Intermediate Guidance}
\label{sec:method}

\subsection{LLM-Generated Guidance}
\label{sec:progressive-oracle-supervision}

In KAME, an asynchronous LLM is introduced to supply candidate responses to the frontend S2S model \cite{kuroki2026kame}.
At inference time, this asynchronous LLM is queried periodically at 2 Hz to generate candidate responses from the partial user inputs.
We refer to this series of updates as the guidance stream (called the ``oracle'' stream in \cite{kuroki2026kame}).

A key difficulty in training the tandem S2S model is that ordinary dialogue data provide user utterances and their target responses, but not the corresponding guidance stream.
In KAME, a simulator LLM is employed to simulate a trace that converges to the target response (see ``LLM-generated'' in Fig.~\ref{fig:guidance-overview}).
At successive points in the user utterance, the simulator generates a possible interpolation between the response estimated from the partial input and the target transcript $y$.
This guidance stream is designed to end with $y$ itself.
With this simulation, the frontend S2S model is optimized to start generating response audio once relevant information is obtained from the guidance stream.
To prepare a training dataset, this procedure has to be repeated for each training dialogue.

\subsection{Randomized Intermediate Guidance}
\label{sec:randomized-intermediate-guidance}
To avoid costly LLM simulation in large-scale corpus preparation, we propose a method that derives the training-time guidance stream directly from the training corpus itself. 
This method is grounded in our hypothesis about the KAME training process: the most critical aspect of KAME fine-tuning is teaching the model to distinguish informative guidance from irrelevant guidance that arises from premature LLM invocation.
In this method, the guidance stream is designed to contain potentially irrelevant guidance texts randomly sampled from other conversation histories alongside the relevant guidance text copied from the target response (see ``Randomized'' in Fig.~\ref{fig:guidance-overview}).

For each response segment, we retain the target transcript $y$ at the last designated target-response update and fill the remaining updates with randomly sampled response texts. 
We form a pool of response texts from the training corpus, deduplicated by token sequence, and exclude all response texts appearing in the current dialogue.
We retain candidates containing between half and twice as many tokens as $y$.
For each segment, we sample the required response texts uniformly without replacement from the filtered pool.

\subsection{Application to Real Conversations}
\label{sec:natural-conversation-guidance}

To use randomized guidance with real conversations, we first prepare
clean single-speaker audio segments and time-aligned transcripts.
We use Silero VAD to detect speech regions in PodcastIndex recordings, then apply the pyannote speaker-diarization-3.1 pipeline \cite{bredin2023pyannote} for speaker diarization on each channel.
We project the diarization results onto the VAD segments and discard segments containing multiple speakers or overlapping speech.
We then enhance each retained segment with Sidon \cite{nakata2026sidon},
filter it using DNSMOS P.835 \cite{reddy2022dnsmos} (OVRL $\geq 3.0$),
and transcribe it with batched faster-whisper large-v3-turbo.
We obtain token-level timestamps using faster-whisper’s
Whisper-based alignment routine \cite{radford2023whisper}. 
We further filter transcriptions using Whisper language
probability ($\geq 0.85$) and average log probability
($\geq -0.4$), and require non-empty transcripts.
After this segmentation and filtering process, we
construct dialogue examples from eligible segment groups associated with the same source recording and source channel and containing exactly two speaker identities. 
We order the selected segments by their original timestamps and concatenate them back-to-back. This preserves the relative chronological order of the retained segments while discarding the original inter-segment gaps.

For each response segment, the recorded audio and its aligned transcript serve as the training targets. 
The same transcript also serves as target-response guidance $y$, while texts sampled from other turns in the training corpus provide randomized intermediate guidance. 
The corpus thus supplies both the training targets and the guidance stream without per-example LLM generation.

%% file: sections/experiments.tex
\section{Experiments}
\label{sec:experiments}

\subsection{Response Quality with Randomized Guidance}
\label{sec:semantic-fidelity-experiment}

We compare four guidance-con\-struc\-tion strategies on the same English synthetic
dialogues: LLM-generated, Sim\-i\-lar\-ity-based, Target-only, and Randomized (ours).

The first strategy, LLM-generated, is our baseline and follows the original KAME construction described in Section~\ref{sec:progressive-oracle-supervision} \cite{kuroki2026kame}. 
The second strategy, Sim\-i\-lar\-ity-based, avoids per-ex\-am\-ple LLM generation while still approximating the backend's semantic progression through retrieval. 
We first derive similarity trajectories from 30 calibration turns by comparing the backend's responses to partial user utterances with its response to each complete utterance, using a frozen MiniLM-based sentence encoder \cite{wang2020minilm}.
For each training segment, we sample one of these trajectories rather than an averaged curve, and retrieve corpus responses whose cosine similarity to the target transcript $y$ approximately follows the piecewise-linear interpolation of the sampled trajectory.
These trajectories need not increase monotonically.
We also examine whether the target response alone provides sufficient guidance for competitive response quality. The third strategy,
Target-only, therefore omits intermediate guidance. 
Finally, Randomized (ours) samples intermediate guidance from other training dialogues without semantic selection (Section~\ref{sec:randomized-intermediate-guidance}). 
This combination is intended to train the frontend to use informative guidance while ignoring irrelevant updates.

For all four strategies, we fine-tune all parameters of the same pretrained Moshi checkpoint for one epoch. We use a global batch size of 32 and learning rates of $2\times10^{-6}$ and $4\times10^{-6}$ for the temporal and depth transformers, respectively.

We evaluate response quality on a spoken MT-Bench subset
\cite{zheng2023judging} comprising 30 questions, each with two turns. All models use GPT-4.1 as the inference-time backend. We transcribe their generated responses with Whisper large-v3 \cite{radford2023whisper} and, following the original MT-Bench evaluation protocol, score the transcripts with a GPT-4 judge on a 1--10 scale. 
For each model, we repeat judging three times with the audio and transcripts held fixed, and report the mean and standard error across the three 60-turn averages.

\input{tables/synthetic-guidance}
\input{tables/natural-conversations}

Table~\ref{tab:synthetic-guidance} shows that Randomized achieves competitive response quality, scoring 6.09 compared with 6.14 for Sim\-i\-lar\-ity-based and 5.54 for LLM-generated (KAME). It also scores higher than Target-only (5.12). 
One possible explanation for the improvement over LLM-generated guidance is that the simulated intermediate guidance in KAME follows a heuristically designed progression toward the target response and may not fully reflect the variability of inference-time backend updates. In contrast, randomized guidance can be viewed as a form of negative sampling: by exposing the frontend to both informative target-response guidance and potentially irrelevant intermediate updates, it may encourage the model to rely on guidance only when it is useful.
Together, these results support randomized guidance as a practical alternative to reconstructing intermediate backend predictions.

\subsection{Training on Real Conversations}
\label{sec:natural-conversation-experiment}

Randomized guidance allows KAME to be trained on a large-scale real dataset without per-example LLM generation. We prepare 3.8k hours of English two-speaker conversations following Section~\ref{sec:natural-conversation-guidance}, approximately $2.8\times$ the audio duration of the 1.36k-hour synthetic corpus used for the synthetic-data KAME baseline.
The central question is whether learning from
these recordings improves conversational timing and delivery while retaining KAME's response-quality advantage over Moshi. We address this question by
comparing the real-data model with pretrained Moshi and KAME (synthetic data), the LLM-generated baseline from Section~\ref{sec:semantic-fidelity-experiment}.

We fine-tune all parameters of the same pretrained Moshi checkpoint as in
Section~\ref{sec:semantic-fidelity-experiment}. With batch size and
learning rates unchanged, we evaluate the 3,000-update checkpoint, denoted KAME (real data, ours).

We assess response quality with MT-Bench and conversational behavior with
three tasks derived from MTR-DuplexBench \cite{he2026mtrduplexbench}: smooth
turn-taking, interruption handling, and pause handling. These tasks examine
whether a model responds after the user finishes speaking, yields when
interrupted, and waits through pauses within a user turn. For each task,
we average success over 200 ten-round dialogues and three generation seeds,
including the interruption task's initial setup round. Both KAME models
receive identical precomputed GPT-4.1 guidance streams from user-transcript
prefixes at 0.5-second intervals, with a fixed simulated backend delay of six frames (0.48~s).
We use local acoustic/ASR-based scoring; these rates are not directly
comparable to published benchmark scores.

To assess conversational delivery, we use Gemini 3.1 Pro (high thinking level) as an audio judge, following prior work on audio-aware judging \cite{chiang2025audiojudges}.
It scores naturalness (1--10) from response-only clips of up to ten seconds, starting at question end, for both turns of 30 MT-Bench questions.
The evaluation is designed to focus on the audible naturalness of the response: no question text, transcripts, or model identities are provided.
The prompt assesses ordinary, unperformed conversational delivery without
requiring polish or expressiveness. Deductions require audible evidence;
answer content, turn-taking, leading silence, and clip-boundary artifacts
are excluded. Clips with insufficient speech are unassessable.
We report mean and standard error across three repetition-level averages
on fixed audio, using the 16 questions with both turns assessable for every
model in every repetition.

The results in Table~\ref{tab:natural-conversations} show that real-data KAME narrows the naturalness gap to Moshi while preserving a substantial response-quality advantage. 
Audio-judge naturalness rises from the synthetic baseline's 4.05 to 4.79, approaching Moshi's 5.29. At the same time, its MT-Bench score of 5.04 remains within 0.50 points of synthetic-data KAME and well above Moshi's 1.96. 
Smooth turn-taking success also increases from 40.62\% to 50.88\%, extending the gains beyond vocal delivery to turn-taking.
Pause handling remains similar, while interruption handling remains a limitation. 
One possible reason is that our training objective does not explicitly target stopping an ongoing response and responding again after the user finishes speaking. 

%% file: tables/synthetic-guidance.tex
\begin{table}[t]
    \centering
    \caption{Response quality of four guidance-construction strategies for KAME on the
    30-question spoken MT-Bench subset.
    Mean $\pm$ standard error over three GPT-4 judge repetitions
    using fixed audio and transcripts.}
    \label{tab:synthetic-guidance}
    \begin{tabular}{llc}
        \toprule
        Guidance strategy & \shortstack[l]{Intermediate\\guidance} & MT-Bench \\
        \midrule
        LLM-generated (KAME) & LLM       & $5.54 \pm 0.03$ \\
        Similarity-based     & Retrieval & $6.14 \pm 0.05$ \\
        Target-only    & None      & $5.12 \pm 0.03$ \\
        Randomized (ours) & Random    & $6.09 \pm 0.05$ \\
        \bottomrule
    \end{tabular}
\end{table}

%% file: tables/natural-conversations.tex
\begin{table*}[t]
    \centering
    \caption{Response quality, conversational behavior, and audio-judge naturalness for three
    models; higher is better. MTR-derived success rates average three
    generation seeds. MT-Bench and naturalness report mean $\pm$ standard error.
    MT-Bench and naturalness use three judge repetitions on fixed
    audio; naturalness uses the common 16-question subset.}
    \label{tab:natural-conversations}
    \begin{tabular}{lccccc}
        \toprule
        & MT-Bench & \multicolumn{3}{c}{MTR-derived success (\%)}
        & Audio-judge naturalness \\
        Model & (1--10) & Smooth & Interruption & Pause & (1--10) \\
        \midrule
        KAME (synthetic data) & $5.54 \pm 0.03$ & 40.62 & 19.28 & 80.20
                         & $4.05 \pm 0.14$ \\
        Moshi            & $1.96 \pm 0.01$ & 36.75 & 23.93 & 74.62
                         & $5.29 \pm 0.22$ \\
        KAME (real data, ours) & $5.04 \pm 0.21$ & 50.88 & 17.77 & 81.08
                         & $4.79 \pm 0.32$ \\
        \bottomrule
    \end{tabular}
\end{table*}

%% file: sections/conclusion.tex
\section{Conclusion}
\label{sec:conclusion}

In this paper, we introduced randomized intermediate guidance, which constructs the missing training-time guidance stream directly from the corpus by combining target-response guidance with randomly sampled response texts, avoiding per-example LLM simulation.
Experiments on synthetic dialogues show that randomized guidance achieves response quality comparable to the LLM-generated and similarity-based baselines.
Applying this approach to 3.8k hours of real conversations yields smoother turn-taking and higher audio-judge naturalness than synthetic-data KAME, while retaining a substantial MT-Bench advantage over Moshi.
Thus, we show that randomized intermediate guidance supports effective KAME training and enables more natural interaction when combined with real conversational data.

%% file: refs.bib
@inproceedings{bredin2023pyannote,
  author = {Herv{\'e} Bredin},
  title = {{pyannote.audio} 2.1 Speaker Diarization Pipeline: Principle,
           Benchmark, and Recipe},
  booktitle = {Proc. Interspeech 2023},
  year = {2023},
  pages = {1983--1987},
  doi = {10.21437/Interspeech.2023-105},
  url = {https://www.isca-archive.org/interspeech_2023/bredin23_interspeech.html}
}

@inproceedings{wang2020minilm,
  author = {Wenhui Wang and Furu Wei and Li Dong and Hangbo Bao and
            Nan Yang and Ming Zhou},
  title = {{MiniLM}: Deep Self-Attention Distillation for Task-Agnostic
           Compression of Pre-Trained Transformers},
  booktitle = {Advances in Neural Information Processing Systems},
  volume = {33},
  year = {2020},
  url = {https://proceedings.neurips.cc/paper/2020/hash/3f5ee243547dee91fbd053c1c4a845aa-Abstract.html}
}

@inproceedings{radford2023whisper,
  author = {Alec Radford and Jong Wook Kim and Tao Xu and Greg Brockman
            and Christine McLeavey and Ilya Sutskever},
  title = {Robust Speech Recognition via Large-Scale Weak Supervision},
  booktitle = {Proceedings of the 40th International Conference on Machine
               Learning},
  series = {Proceedings of Machine Learning Research},
  volume = {202},
  pages = {28492--28518},
  publisher = {PMLR},
  year = {2023},
  url = {https://proceedings.mlr.press/v202/radford23a.html}
}

@inproceedings{zheng2023judging,
  author = {Lianmin Zheng and Wei-Lin Chiang and Ying Sheng and Siyuan Zhuang
            and Zhanghao Wu and Yonghao Zhuang and Zi Lin and Zhuohan Li and
            Dacheng Li and Eric P. Xing and Hao Zhang and Joseph E. Gonzalez
            and Ion Stoica},
  title = {Judging {LLM}-as-a-Judge with {MT-Bench} and {Chatbot Arena}},
  booktitle = {Advances in Neural Information Processing Systems},
  volume = {36},
  year = {2023},
  url = {https://proceedings.neurips.cc/paper_files/paper/2023/hash/91f18a1287b398d378ef22505bf41832-Abstract-Datasets_and_Benchmarks.html}
}

@article{defossez2024moshi,
  author = {Alexandre D{\'e}fossez and Laurent Mazar{\'e} and Manu Orsini and
            Am{\'e}lie Royer and Patrick P{\'e}rez and Herv{\'e} J{\'e}gou and
            Edouard Grave and Neil Zeghidour},
  title = {{Moshi}: A Speech-Text Foundation Model for Real-Time Dialogue},
  journal = {arXiv preprint arXiv:2410.00037},
  year = {2024},
  doi = {10.48550/arXiv.2410.00037},
  url = {https://arxiv.org/abs/2410.00037}
}

@inproceedings{kuroki2026kame,
  author = {So Kuroki and Yotaro Kubo and Takuya Akiba and Yujin Tang},
  title = {{KAME}: Tandem Architecture for Enhancing Knowledge in Real-Time
           Speech-to-Speech Conversational {AI}},
  booktitle = {Proc. IEEE International Conference on Acoustics, Speech and
               Signal Processing (ICASSP)},
  year = {2026},
  url = {https://www.cmsworkshops.com/ICASSP2026/view_paper.php?PaperNum=11317}
}

@article{ohashi2026interactivity,
  author = {Atsumoto Ohashi and Neil Zeghidour and Alexandre D{\'e}fossez
            and Eugene Kharitonov},
  title = {Multi-Faceted Interactivity Alignment in Full-Duplex Speech Models},
  journal = {arXiv preprint arXiv:2606.11167},
  year = {2026},
  doi = {10.48550/arXiv.2606.11167},
  url = {https://arxiv.org/abs/2606.11167}
}

@article{he2026conversationalvoice,
  author = {Richard Yucheng He and Baodong Cao and Chen Xu and Yihang Liu
            and Tairan Chen},
  title = {{ConversationalVoice}: Full-Duplex Speech Data from Real
           Conversations through Source-Faithful Reconstruction and
           Conversation-Grounded Expansion},
  journal = {arXiv preprint arXiv:2609.08147},
  year = {2026},
  doi = {10.48550/arXiv.2609.08147},
  url = {https://arxiv.org/abs/2609.08147}
}

@article{nakata2026duplexchat,
  author = {Wataru Nakata and Yuki Saito and Hiroshi Saruwatari},
  title = {{DuplexChat}: Constructing Speaker-Separated Full-Duplex Dialogue
           Speech at Scale for Spoken Dialogue Language Modeling},
  journal = {arXiv preprint arXiv:2607.04941},
  year = {2026},
  doi = {10.48550/arXiv.2607.04941},
  url = {https://arxiv.org/abs/2607.04941}
}

@article{chien2026moshirag,
  author = {Chung-Ming Chien and Manu Orsini and Eugene Kharitonov and
            Neil Zeghidour and Karen Livescu and Alexandre D{\'e}fossez},
  title = {{MoshiRAG}: Asynchronous Knowledge Retrieval for Full-Duplex
           Speech Language Models},
  journal = {arXiv preprint arXiv:2604.12928},
  year = {2026},
  doi = {10.48550/arXiv.2604.12928},
  url = {https://arxiv.org/abs/2604.12928}
}

@article{srinivas2026thinking,
  author = {Vidya Srinivas and Zachary Englhardt and Vikram Iyer and
            Shwetak Patel},
  title = {Thinking While Speaking: Inference-Time Knowledge Transfer for
           Responsive and Intelligent Conversational Voice Agents},
  journal = {arXiv preprint arXiv:2511.07397},
  year = {2026},
  doi = {10.48550/arXiv.2511.07397},
  url = {https://arxiv.org/abs/2511.07397}
}

@inproceedings{he2026mtrduplexbench,
  author = {He, Zhang and Cui, Wenqian and Xu, Haoning and Li, Xiao-Hui and
            Zhu, Lei and Bai, Haoli and Shaohua, Ma and King, Irwin},
  title = {{MTR-DuplexBench}: Towards a Comprehensive Evaluation of
           Multi-Round Conversations for Full-Duplex Speech Language Models},
  booktitle = {Findings of the Association for Computational Linguistics:
               ACL 2026},
  year = {2026},
  pages = {5334--5351},
  publisher = {Association for Computational Linguistics},
  address = {San Diego, California, United States},
  doi = {10.18653/v1/2026.findings-acl.263},
  url = {https://aclanthology.org/2026.findings-acl.263/}
}

@inproceedings{chiang2025audiojudges,
  author = {Cheng-Han Chiang and Xiaofei Wang and Chung-Ching Lin and
            Kevin Lin and Linjie Li and Radu Kopetz and Yao Qian and
            Zhendong Wang and Zhengyuan Yang and {Hung-yi} Lee and Lijuan Wang},
  title = {Audio-Aware Large Language Models as Judges for Speaking Styles},
  booktitle = {Findings of the Association for Computational Linguistics:
               EMNLP 2025},
  year = {2025},
  pages = {467--480},
  publisher = {Association for Computational Linguistics},
  address = {Suzhou, China},
  doi = {10.18653/v1/2025.findings-emnlp.25},
  url = {https://aclanthology.org/2025.findings-emnlp.25/}
}

@inproceedings{nakata2026sidon,
  title={Sidon: Fast and robust open-source multilingual speech restoration for large-scale dataset cleansing},
  author={Nakata, Wataru and Saito, Yuki and Ueda, Yota and Saruwatari, Hiroshi},
  booktitle={Proc. IEEE International Conference on Acoustics, Speech and Signal Processing (ICASSP)},
  pages={17617--17621},
  year={2026},
  organization={IEEE}
}

@inproceedings{reddy2022dnsmos,
  title={{DNSMOS P.835}: A non-intrusive perceptual objective speech quality metric to evaluate noise suppressors},
  author={Reddy, Chandan KA and Gopal, Vishak and Cutler, Ross},
  booktitle={Proc. IEEE International Conference on Acoustics, Speech and Signal Processing (ICASSP)},
  pages={886--890},
  year={2022},
  organization={IEEE}
}
